\documentclass{article}
\usepackage{multirow}
\usepackage{amssymb}% http://ctan.org/pkg/amssymb
\usepackage{pifont}% http://ctan.org/pkg/pifont
\usepackage{ijcai21}
\usepackage{times}
\usepackage{soul}
\usepackage{url}
\usepackage[hidelinks]{hyperref}
\usepackage[utf8]{inputenc}
\usepackage[small]{caption}
\usepackage{graphicx}
\usepackage{amsmath}
\usepackage{amsthm}
\usepackage{booktabs}
\usepackage{algorithm}
\usepackage{algorithmic}
\usepackage{CJKutf8}
\usepackage{color}

\title{PIANO: A Parametric Hand Bone Model from Magnetic Resonance Imaging}

\author{
Yuwei Li$^{1,2,3}$\and
Minye Wu$^{1,2,3}$\and
Yuyao Zhang$^1$\and
Lan Xu$^{1}$\And
Jingyi Yu$^{1}$\\
\affiliations
$^1$Shanghai Engineering Research Center of Intelligent Vision and Imaging, School of Information Science and Technology, ShanghaiTech
University\\
$^2$Shanghai Institute of Microsystem and Information Technology,
Chinese Academy of Sciences
\\
$^3$University of Chinese Academy of Sciences\\
\emails
\{liyw, wumy, zhangyy8, xulan1, yujingyi\}@shanghaitech.edu.cn
}

\begin{document}

\maketitle

\begin{abstract}
Hand modeling is critical for immersive VR/AR, action understanding, or human healthcare.
Existing parametric models account only for hand shape, pose, or texture, without modeling the anatomical attributes like bone, which is essential for realistic hand biomechanics analysis.
In this paper, we present PIANO, the first parametric bone model of human hands from MRI data.
Our PIANO model is biologically correct, simple to animate, and differentiable, achieving more anatomically precise modeling of the inner hand kinematic structure in a data-driven manner than the traditional hand models based on the outer surface only.
Furthermore, our PIANO model can be applied in neural network layers to enable training with a fine-grained semantic loss, which opens up the new task of data-driven fine-grained hand bone anatomic and semantic understanding from MRI or even RGB images.
We make our model publicly available.
\end{abstract}

\newcommand{\myparagraph}[1]{\vspace{0.1em}\noindent\textbf{#1}}

\begin{CJK}{UTF8}{gbsn}
    \section{Introduction}
% 1. Top view
Hands are essential for humans to interact with the world and hand modeling enables numerous applications for immersive VR/AR, action analysis and healthcare, etc.
However, reconstructing realistic human hands in a biologically and anatomically plausible manner remains challenging and has received substantive attention in both the artificial intelligence and computer vision communities.

\begin{figure}[tbp] 
	\centering 
	\includegraphics[width=\linewidth]{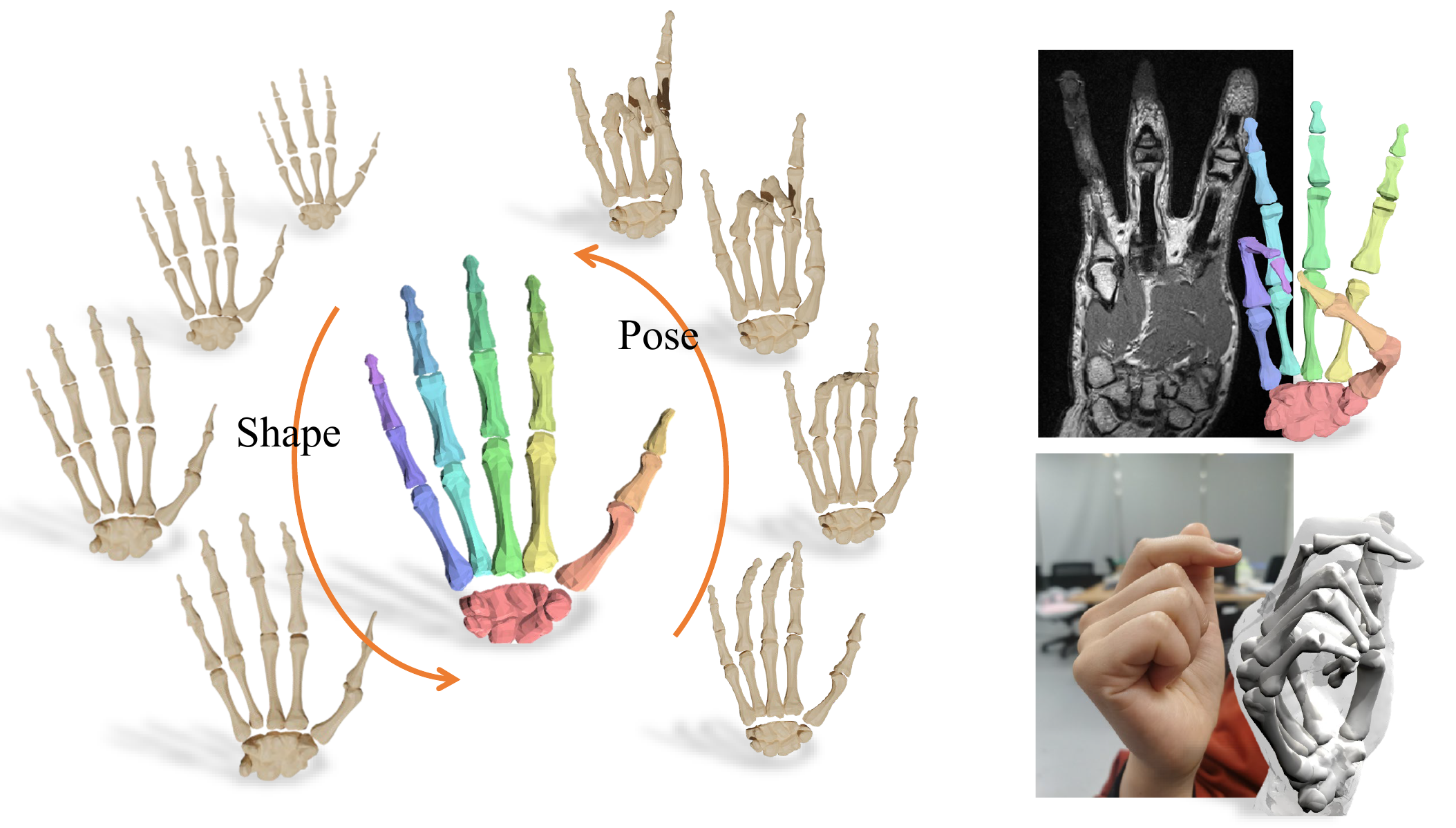} 
	\caption{We present the first statistical internal hand bone model. Our model enables anatomically and physically precise modeling of the hand kinematic structure of different individuals, opening up the study of data-driven hand bone anatomic and semantic fine-grained understanding from MRI or even RGB images.} 
	\label{fig:fig_1_teaser} 
\end{figure}

In recent years, there has been significant progress in hand pose, shape, geometry
~\cite{wang2020rgb2hands,smith2020constraining}, and even texture estimation~\cite{qian2020html} in a data-driven manner with the corresponding parametric models~\cite{romero2017embodied,moon2020deephandmesh}.
In contrast, there is little work that addresses the parametric modeling and data-driven estimation of human hand anatomy like bone and muscle.
However, such fine-grained hand anatomical attribute modeling is important for understanding the complex hand biomechanics and producing realistic and anatomically precise virtual hands.
% 2.2 MRI/CT --> anatomic modeling --> no parametric model yet, let alone data-driven understanding 
On the other hand, modern 3D medical imaging techniques like Magnetic Resonance Imaging (MRI) and Computed Tomography (CT) are widely used for treating hand injuries clinically ~\cite{kapandji2007physiology} or modeling the function of healthy hands~\cite{MiyataFingerJoint}.
% no parametric model, no data-driven possibilities
However, the above-mentioned modeling methods focus on the analysis of only a single individual. 
% CT-> healthy issue
Moreover, those CT-based methods~\cite{kurihara2004modeling,Yang_2020} suffer from ethical issues due to the unacceptably high radiation doses during the scanning of various individuals under various hand poses for building a high-quality parametric hand bone model.
% MRI
The recent MRI-based method~\cite{wang2019hand} adopts a molding procedure to build an anatomy-correct hand bone rig of a target performer. However, it suffers from time-consuming and user-specified bone segmentation operations, which is impractical to apply to various individuals for building a parametric model.

% 3. Our key idea
To address the gaps above, we present the first statistical hand bone model from MRI data called \textit{PIANO} for \textit{ParametrIc hANd bOne model}, which enables anatomically and physically precise modeling of the hand kinematic structure of different individuals (see Fig.~\ref{fig:fig_1_teaser}).
Our model is biologically correct, simple to animate, and differentiable, opening up the study of data-driven hand bone anatomic and semantic fine-grained understanding from MRI or even RGB images at a level of detail not previously possible.

% 4. Our technical pipeline
% 4.1 Dataset
To this end, we first collect a new large-scale dataset of high-quality MRI hand scans using a clinical 3T scanner and an efficient simplified molding procedure from \cite{wang2019hand},
which consists of 200 scans of 35 subjects in up to 50 poses with rich and accurate 3D joint and bone segmentation annotations from experienced radiologists.  
% 4.2 Model
Then, we use this data as well as an artist-rigged hand bone template mesh with 20 separated bones to train the statistical hand bone model PIANO similar to the outer human body surface model SMPL~\cite{loper2015smpl} and hand model MANO~\cite{romero2017embodied}.
Due to the inherent per-bone rigidness in PIANO, we omit the pose blend shape and adopt a pre-defined blending weight. 
Thus, PIANO inherently factors the per-bone anatomic geometric changes of each individual and has analogous components to those in SMPL and MANO: a hand bone template mesh with kinematic tree, shape blend shape, and a joint regressor.
Especially, to build PIANO in an anatomically precise manner, we introduce a novel coarse-to-fine multi-stage registration approach and a bone-aware model learning strategy, which jointly considers the accurate joint and bone segmentation annotations provided by radiologists.
% 4.3 Application: proof-of-concept
Our new parametric hand bone model allows obtaining more anatomically precise personalized rigging than the traditional hand models based on outer surface only. 
Furthermore, we introduce a novel neural network based on PIANO to produce fine-grained and biologically correct hand bone segmentation from an MRI scan or even a single RGB image input in an analysis-by-synthesis fashion.
It enables a fine-grained semantic loss and opens up a new task for data-driven internal hand bone modeling unseen before.
% 5. Our technical contributions
To summarize, our main technical contributions include:
\begin{itemize} 
	\setlength\itemsep{0em}
	\item We present the first parametric hand bone model, PIANO, based on a new MRI hand dataset, which enables anatomically precise internal kinematic understanding of human hands in a data-driven manner.
	
	\item We propose a novel multi-stage registration approach as well as a bone-aware model learning strategy, which utilizes the joint and bone segmentation annotations in MRI data provided by radiologists to build PIANO. 
	
	\item We present a proof-of-concept neural network design that enables data-driven fine-grained hand bone segmentation from an MRI scan or RGB image.
	
	\item We make available our PIANO model and the new dataset of 200 MRI hand scans of 35 subjects with various poses and rich annotations\footnote{\url{https://reyuwei.github.io/proj/piano.html}}.
\end{itemize}

    \section{Related Work}

\noindent\textbf{Parametric Modeling.}
Using parametric model for human performance capture on body~\cite{loper2015smpl} and hand~\cite{romero2017embodied,smith2020constraining} have been widely studied for the past decades. 
% little on inner structure
Most of them focus on outer information like skin surface and texture~\cite{qian2020html,moon2020deephandmesh,van2018articulating}.
\textcolor{black}{\cite{mirakhorlo2018musculoskeletal} proposed a comprehensive biomechanical hand model based on detailed measurments from a hand specimen, yet it is not differentiable and can not be embedded in deep learning frameworks. 
}
% carpal pm model for segmentation
\cite{wrist_pmodel_16} proposed a statistical wrist shape and bone model for automatic carpal bone segmentation.
On this basis, we utilize medical imaging techniques and build a parametric hand bone model which is anatomically correct, physically precise and differentiable.

\noindent\textbf{MRI/CT hand analysis.}
Common medical imaging techniques that are widely used for hand treatment include Computed Tomography (CT) and Magnetic Resonance Imaging (MRI)~\cite{kapandji2007physiology}.
% CT - radiation
Pioneering work of \cite{kurihara2004modeling} and \cite{marai20033d} have used CT scans to acquire hand bone skeleton and carpal movements. 
\textcolor{black}{\cite{martin2009automatic} proposed automatic articulated registration of hands out of X-ray images.}
However, CT and X-ray emits ionizing radiation, while MRI is safer and provides more clear contrast between soft body tissue. 
% \cite{rhee2007soft,MiyataFingerJoint,Shimizu2010ConstructingM3} 
\cite{rhee2007soft,MiyataFingerJoint} used MRI for understanding the hand anatomy and biomechanics on a small set of poses. Recently, 
% MRI - molding
\cite{wang2019hand} adopts a molding procedure for long time MRI scan, enabling anatomical hand bone analysis with complex poses. 
% MRI - single individual
However, suffering from time-consuming and user-specified operations, these methods only focus on single individuals. 
% Our
Comparably, we employ a simplified molding procedure and a semi-automatic annotation method to build a large-scale dataset of high-quality MRI hand scans with precise annotation on bone and joints.
    \section{Parametric Hand Bone Model} \label{sec:modeling}
In this work, our goal is to build a parametric internal bone model of human hands, which is anatomically correct, simple to animate and differentiable. % a parametric internal hand bone model
First, we build a new MRI hand dataset via an efficient molding procedure, which captures precise anatomical structure of human hands with rich annotations provided by experienced radiologists (Sec.~\ref{subsec:data}).
Then, similar to outer surface modeling, we train our PIANO model to obtain the hand bone template mesh with kinematic tree, shape blend shape and joint regressor by iterating between model registration and model parameter learning in a flip-flop manner.
Our detailed model formulation is provided in Sec.~\ref{subsec:model_define}, followed by the illustration of our novel multi-stage model registration (Sec.~\ref{subsec:registration}) and our bone-aware parameter learning scheme (Sec.~\ref{subsec:training}), so as to utilize the rich annotations in MRI data provided by radiologists.

\subsection{Hand MRI Data Collection} \label{subsec:data}
\begin{figure}
	\centering
	\includegraphics[width=\linewidth]{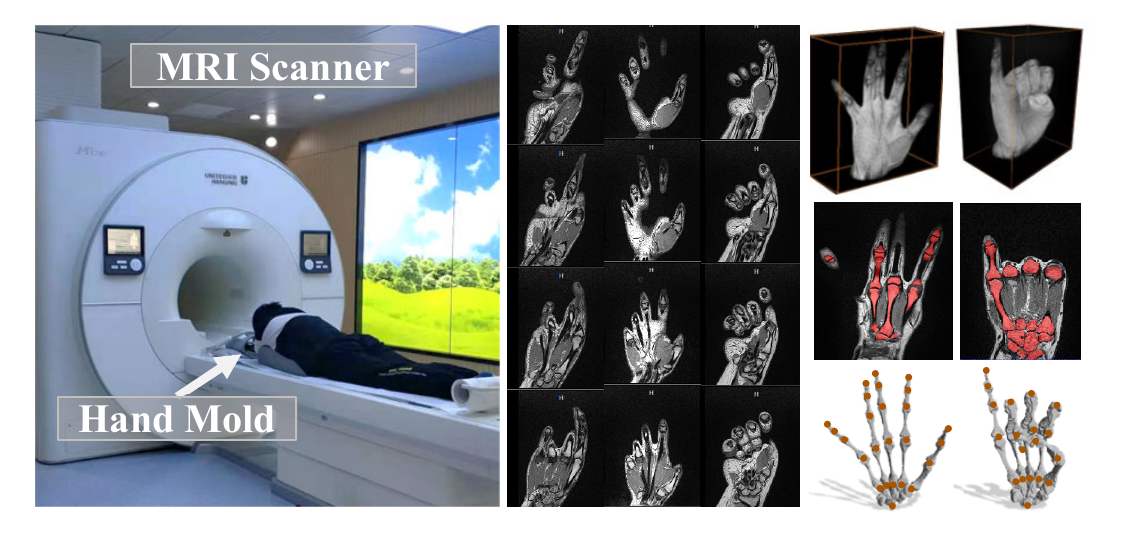}
	\caption{Left: MRI scanner and participant. Middle: MRI raw slices. Right: Dataset samples. From top to bottom are MRI volume, binary bone mask and bone mesh overlayed with 25 precise joint annotation. }
	\label{fig:dataset}
\end{figure}

Magnetic Resonance Imaging (MRI) scanning is performed to capture the anatomical structure under each hand pose.
In all, a total of 200 hand volumetric images are captured with 50 different hand posture of 35 individual subjects.
The scanning is performed with a 3.0 Tesla MR scanner of United Imaging (uMR780, Shanghai) with spatial resolution $0.4\times0.4\times1~mm^3$ and T1-weighted sequence. 
Since MRI is a motion-sensitive imaging technique, during scanning, we adopt a simplified molding procedure to stabilize hand poses similar as that in \cite{wang2019hand}. 
Briefly, we put hands in a liquid solution which would solidify and encase the hand firmly in less than 3 minutes. We then scan the hand with the mold on for 10 minutes. 

% MRI Annotation
The MRI volume is a regular volumetric grid of scalar values representing tissue mass density. To analyze anatomical structure, bone and joint location must be annotated. However, this task is highly cumbersome and requires strong domain knowledge.
\cite{wang2019hand} proposed an interactive method that takes 3 hours to segment one volume. However, to label all 200 scans in our dataset would take up to 600 hours.
To this end, we adopt a semi-automated human-in-the-loop procedure to label all volumes with a deep neural network. By verifying network predictions and annotating miss labeled samples in an iterative procedure, we only acquire one-third of volume annotations with a moderate manual effort by experienced radiologists. Dataset samples are shown in Fig.\ref{fig:dataset}.

\subsection{Model Formulation} \label{subsec:model_define}
To learn an anatomically correct and simple-to-animate statistical model from our MRI dataset, we follow the outer surface model SMPL~\cite{loper2015smpl} and MANO~\cite{romero2017embodied} to formulate our PIANO model into analogous components, including a hand bone template mesh with kinematic tree, shape blend shape and a joint regressor.

\begin{figure}[t]
	\centering
	\includegraphics[width=\linewidth]{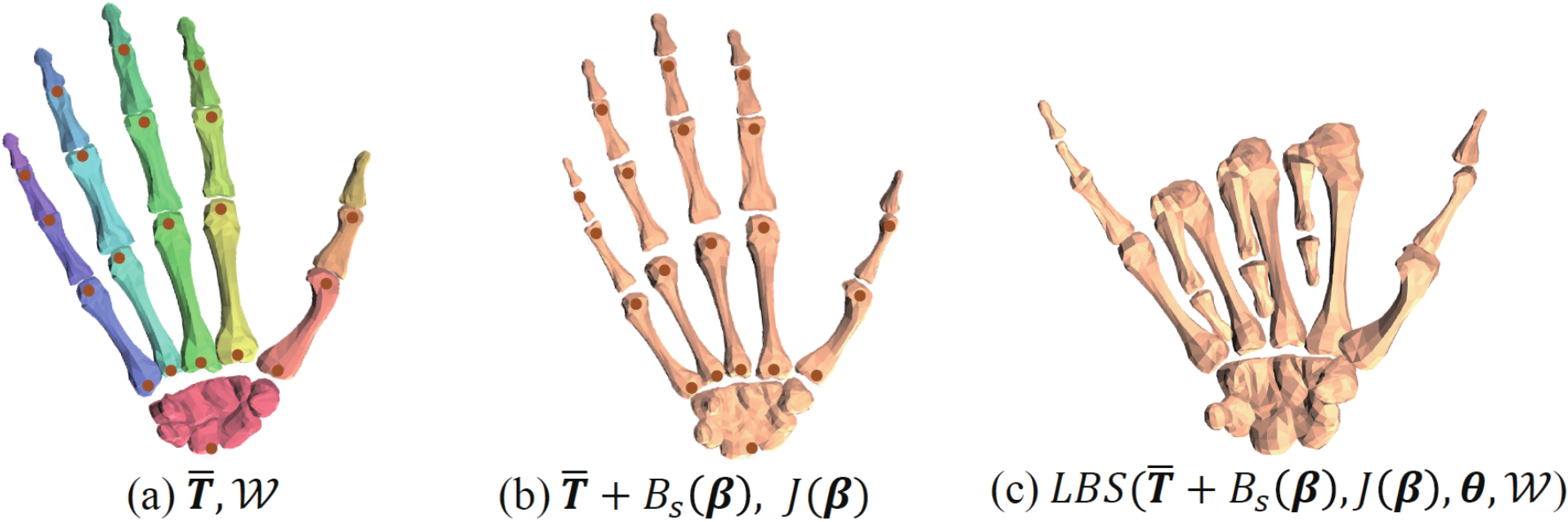}
	\caption{PIANO Model. (a) Template bone mesh with skinning weight indicated by color and joints. (b) With subject-specific shape blend shape. (c) Reposed with linear blend skinning. }
	\label{fig:joint_anatomy}
\end{figure}

As illustrated in Fig.~\ref{fig:joint_anatomy}, we utilize an artist-rigged hand bone template mesh consisting of 20 separated semantic bone meshes with 3345 vertices, 6610 faces and 19 joints plus the global orientation.
Then, our PIANO model defines a function $\mathcal{M}$ of shape $\boldsymbol{\beta}$ and pose $\boldsymbol{\theta}$:
\begin{equation}
\label{eq:modeling:MRIHand}
\mathcal{M}(\boldsymbol{\beta}, \boldsymbol{\theta}) = LBS(\bar{T} + B_S(\boldsymbol{\beta}), J(\boldsymbol{\beta}), \boldsymbol{\theta}, \mathcal{W}).
\end{equation}
Here, $LBS(\cdot)$ denotes the Linear Blend Skinning (LBS) function applied to the warped bone mesh model $T= \bar{T} + B_S(\boldsymbol{\beta})$, skinning weight $\mathcal{W}$ and joint rotation $\boldsymbol{\theta}$.
To model inherent per-bone rigidness, we adopt a fixed blending weight pre-defined by bone label intuitively and omit the pose blend shape.
Thus, we utilize the canonical template mesh $\bar{T} \in \mathbf{R}^{3N}$ of $N = 3345$ vertices under zero pose and shape, as well as the shape blend shape $B_S(\boldsymbol{\beta})$ to factor the per-bone anatomic geometric changes of each individual.
% Shape-blend-shape
In PIANO, we formulate the linear shape blend shape function as $B_S(\boldsymbol{\beta};\mathcal{S}) = \mathcal{S}\boldsymbol{\beta}$ to enable the base shape to vary with identity, where $\boldsymbol{\beta}$ is the shape parameters and matrix $\mathcal{S} \in \mathbf{R}^{3N\times |\boldsymbol{\beta}|}$ represents orthonormal principal components of shape displacements, which is learned from registered training meshes. 
% Joint regressor.
Besides, in PIANO we formulate the joint regressor $J(\boldsymbol{\beta})=\mathcal{J}(\bar{T}+B_S(\boldsymbol{\beta}))$ to map the various shapes to the corresponding anatomically precise joint positions of human hands, where $\mathcal{J} \in \mathbf{R}^{3K\times |\boldsymbol{\beta}|}$ is the learned regression matrix that transforms rest pose vertices into rotation joints; $K=25$ contains 19 rotation joints, 5 finger tips and 1 wrist keypoint, matching our dataset annotation.

Our final PIANO model is $\mathcal{M}(\boldsymbol{\beta}, \boldsymbol{\theta}; \Phi): \mathbf{R}^{|\boldsymbol{\beta}|\times|\boldsymbol{\theta}|} \mapsto \mathbf{R}^{3N}$ which maps the shape and pose parameters to  bone vertices of various individuals ($|\boldsymbol{\beta}| = 10$ and $|\boldsymbol{\theta}| = 57$ in our setting). 
The model parameters $\Phi = \{\bar{T}, \mathcal{S}, \mathcal{J}\}$ includes the canonical template, shape blend shape and joint regressor.

\subsection{Model Registration}\label{subsec:registration}
Here, we propose a multi-stage registration scheme to align the bone template to all the MRI scans, bringing them into correspondence.
Our scheme encodes the internal structure information in a coarse-to-fine manner to fully utilize the anatomically precise joint and bone segmentation annotations in our MRI dataset.
% Let $J^S_l$ denote the annotated 3D location of the $l$-th joint and 
With the joint annotations and bone segmentation, we utilize the radial basis functions (RBF) interpolation~\cite{rhee2007soft} to assign per-bone label of each joint and generate fine-grained segmentation of the bone volume.
Then, we extract both the iso-surface of internal bones $b^S$ and the per-bone 3D geometry $b^S_l$ of the $l$-th joint/bone by applying the Marching Cubes~\cite{lorensen1987marching}.
%  to the corresponding bone segmentation annotations.

\noindent\textbf{Coarse Registration.} We first align the warped joints of the hand bone template mesh to the joint annotations by minimizing the per-joint L2 error and solving the inverse kinematics to obtain the initial pose and shape parameters. 
Since the model parameters $\Phi$ are unknown during the first iteration of model registration and parameter learning, we downgrade our model from Eq.~\ref{eq:modeling:MRIHand} into $\mathcal{M'}(\boldsymbol{\theta}) = LBS(\bar{T}, \bar{J}, \boldsymbol{\theta}, \mathcal{W})$, where we discard shape blend shape function and replace the subject specific parameters \{$T, J$\} with a general template \{$\bar{T}, \bar{J}$\}.

\begin{figure}[t]
	\centering
	\includegraphics[width=0.89\linewidth]{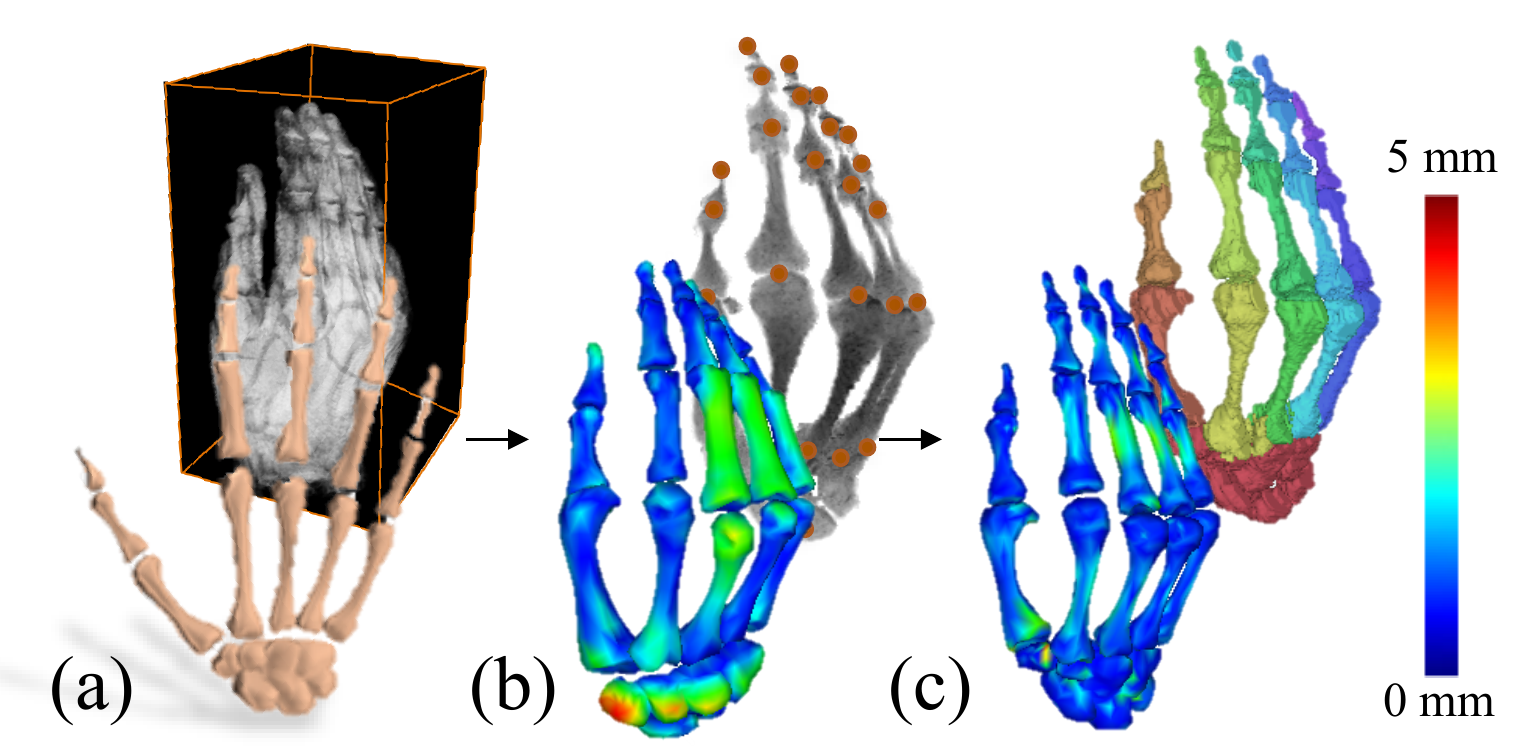}
	\caption{Result of our multistage registration. (a) MRI volume and PIANO template. (b) Coarse stage registration to joint annotation. (c) Fine stage registration to fine-grained bone mesh. We show a color coded Hausdorff distance between registration and annotation. }
	\label{fig:multistage_registration}
\end{figure}

\noindent\textbf{Fine Registration.} We further utilize the internal local and global structural information to perform per-bone non-rigid alignment at a finer scale.
To this end, for the $l$-th bone, we adopt the embedded deformation~\cite{xu2019flyfusion} to register from the initial geometry $b^R_l$ after coarse registration to the annotated geometry $b^S_l$ and $b^S$ from segmentation label.
Let $G$ denote the non-rigid motion field and $ED(\cdot;G)$ denote the embedded deformation function.
%  and please refer to \cite{xu2019flyfusion} for more details.
%
Then, our per-bone alignment is formulated as:
\begin{equation}
\boldsymbol{E}_{\mathrm{fine}}(G) = \boldsymbol{E}_{\mathrm{data}}(G) + \lambda_{\mathrm{reg}} \boldsymbol{E}_{\mathrm{reg}}(G).
\end{equation}
Here, the data term enforces alignment of the warped bone geometry to both current target bone segment $b^S_l$ and the global bone structure $b^S$ to enable robust alignment:
	
\begin{equation}
\label{eq:modeling:registration}
\boldsymbol{E}_{\mathrm{data}}(G) = \sum_{(v,u)\in\mathcal{C}}|| ED(v) - u ||^2_2 + \sum_{(v,u)\in\mathcal{P}}|| ED(v) - u ||^2_2,
\end{equation}
where $\mathcal{C}$ and $\mathcal{P}$ denote the the set of correspondences from $b^R_l$ to $b^S_l$ and $b^S$, respectively, in an Iterative Closest Point (ICP) manner.
To prevent over-fitting to the noise of annotation, we adopt the same locally as-rigid-as-possible motion regularization $\lambda_{\mathrm{reg}}$ and $\boldsymbol{E}_{\mathrm{reg}}$ from \cite{xu2019flyfusion}.	
Fig.~\ref{fig:multistage_registration} shows the intermediate alignment results and demonstrates the effectiveness of our registration scheme. 
After registration, we obtain per-bone aligned meshes with global consistent topology for the following model parameter learning process.

\subsection{Parameter Training} \label{subsec:training}
To train our PIANO model parameters $\Phi = \{\bar{T}, \mathcal{S}, \mathcal{J}\}$, similar to previous work
~\cite{loper2015smpl,romero2017embodied}, 
we adopt a bone-aware training strategy to model pose and shape separately.
Let $b_{i,j,l}$ denote the registered $l$-th bone geometry of the $i$-th subject in the $j$-th hand pose and $J^*_{i,j}$ denote the corresponding annotated 3D joint. 
Then, we optimize the pose parameters $\boldsymbol{\theta}_{i,j}$ that transforms a general template $\bar{T}$ to a specific bone model $b_{i,j}$ with the following energy function:
\begin{equation}
    \boldsymbol{E}_{\mathrm{pose}}(\boldsymbol{\theta}_{i,j}) = \lambda_e \boldsymbol{E}_{\mathrm{edge}}(\boldsymbol{\theta}_{i,j}) + \lambda_j \boldsymbol{E}_{\mathrm{joint}}(\boldsymbol{\theta}_{i,j}).
\end{equation}
Here, similar to SMPL, the edge term measures the edge length difference between two models:
\begin{equation}
   \boldsymbol{E}_{\mathrm{edge}}(\boldsymbol{\theta}_{i,j}) = \sum_{ij}  \sum_l ||\mathcal{E}(LBS(\bar{T}, \bar{J}, \boldsymbol{\theta}_{i,j}, \mathcal{W})_l) - \mathcal{E}(b_{i,j,l})||^2,
\end{equation}
where $\mathcal{E}(\cdot)$ represents the geometry edges. 
Such edge difference enables good estimation of pose without knowing the subject specific shape. 
Then, we add a joint term based on the anatomically precise joint annotation in our dataset:
\begin{equation}
    \boldsymbol{E}_{\mathrm{joint}}(\boldsymbol{\theta}_{i,j}) = \sum_{ij}  ||\bar{J}_{i,j} - {J}^*_{i,j}||^2_2,
\end{equation}
where $\bar{J}_{i,j}$ is the joint position after LBS with the pose $\boldsymbol{\theta}_{i,j}$. 

% step 2
%
Next, we fix the above pose parameters and optimize the subject template $\hat{T}_i$ and joint $\hat{J}_i$ iteratively with the following vertex difference defined on semantic bone segments:
\begin{equation}
\boldsymbol{E}_{\mathrm{vert}}(\hat{T}_i, \hat{J}_i) = \sum_{ij}\sum_l|| LBS(\hat{T}_i, \hat{J}_i, \boldsymbol{\theta}_{i,j}, \mathcal{W})_l - b_{i,j,l}||^2, 
\end{equation}
which enforces the posed subject template to match with the aligned bone meshes. 
We then run principal component analysis (PCA) on $\hat{T}_i$ to obtain shape space parameters $\{\bar{T}, \mathcal{S}\}$, where $\bar{T}$ is the mean shape of $\hat{T}_i$ and $\mathcal{S}$ is a matrix composed of principal component vectors.
% Joints
Also, given subject template $\hat{J}^i$ and $\hat{T}^i$ in rest pose, we further compute the joint regression matrix $\mathcal{J}$ by minimizing $||\mathcal{J}\hat{T}_i - \hat{J}_i||_2^2$.

% % implementation details
To build PIANO, we set $\lambda_{reg}$, $\lambda_e$ and $\lambda_j$ to 5.0, 1.0 and 100.0 respectively and iterate through the whole process of registration and parameter training with Limited-memory BFGS optimizer and PyTorch auto differentiation.

    \begin{figure}
	\centering
	\includegraphics[width=\linewidth]{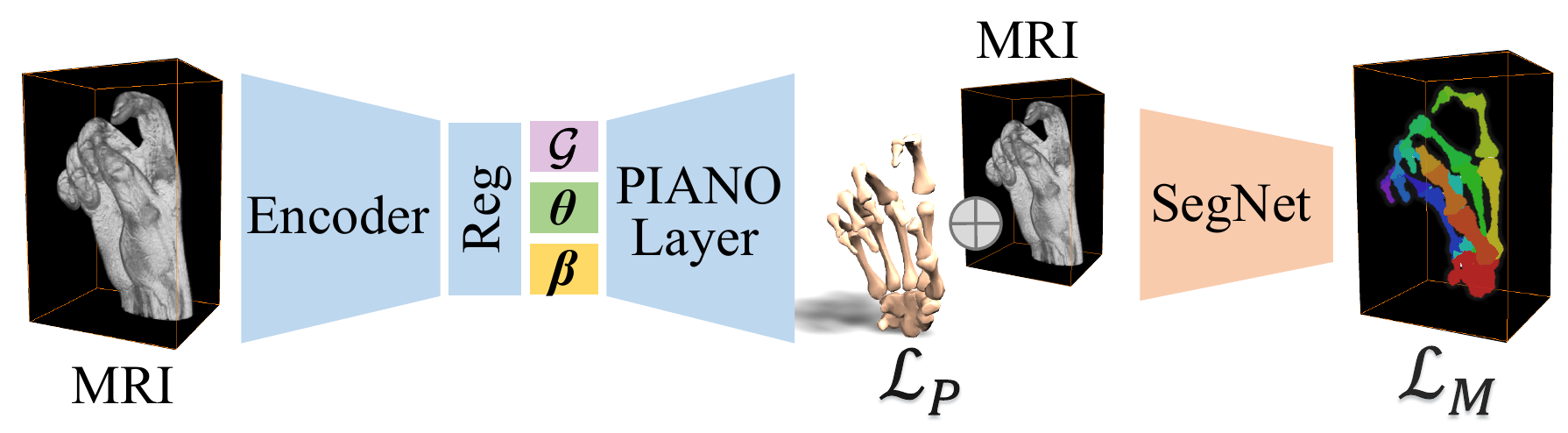}
	\caption{Fine-grained MRI bone segmentation network structure.}
	\label{fig:mri_net}
\end{figure}

\section{Fine-grained Semantic Bone Analysis}
\label{sec:mm}
We now demonstrate the use cases of our statistical model in hand bone anatomic and semantic understanding. We consider the usage of PIANO as a differentiable layer enabling end-to-end MRI fine-grained bone segmentation and joint estimation. Furthermore, we show hand anatomic understanding from a single monocular RGB image.

\noindent\textbf{PIANO Layer.}
We integrate PIANO as a differentiable layer that takes $\boldsymbol{\theta}$ and $\boldsymbol{\beta}$ as input and outputs a triangulated semantic bone mesh and hand joints. 
When training with supervision on PIANO parameters, we define the parameter loss $\mathcal{L}_{pm}$ on predicted shape and pose $ [\hat{\boldsymbol{\beta}}, \hat{\boldsymbol{\theta}}]$ as:
\begin{equation}
    \mathcal{L}_{pm} = || [\hat{\boldsymbol{\beta}}, \hat{\boldsymbol{\theta}}] - [\boldsymbol{\beta}^*, \boldsymbol{\theta}^*] ||^2_2, 
\end{equation}
where  $[\boldsymbol{\beta}^*, \boldsymbol{\theta}^*]$ denotes ground truth.
Alternatively, for training with sparse joint annotations, we define a joint loss as:
\begin{equation}
    \mathcal{L}_J = ||\hat{J} - J^*||^2_2.
\end{equation}
Here, $\hat{J}$ denotes the estimated joints and $J^*$ is the 3D joint annotation,
The resulting PIANO loss $\mathcal{L}_P$ is thus defined as
$\mathcal{L}_P = \mathcal{L}_{pm} + \mathcal{L}_J$.

\begin{figure}[t]
    \centering
    \includegraphics[width=\linewidth]{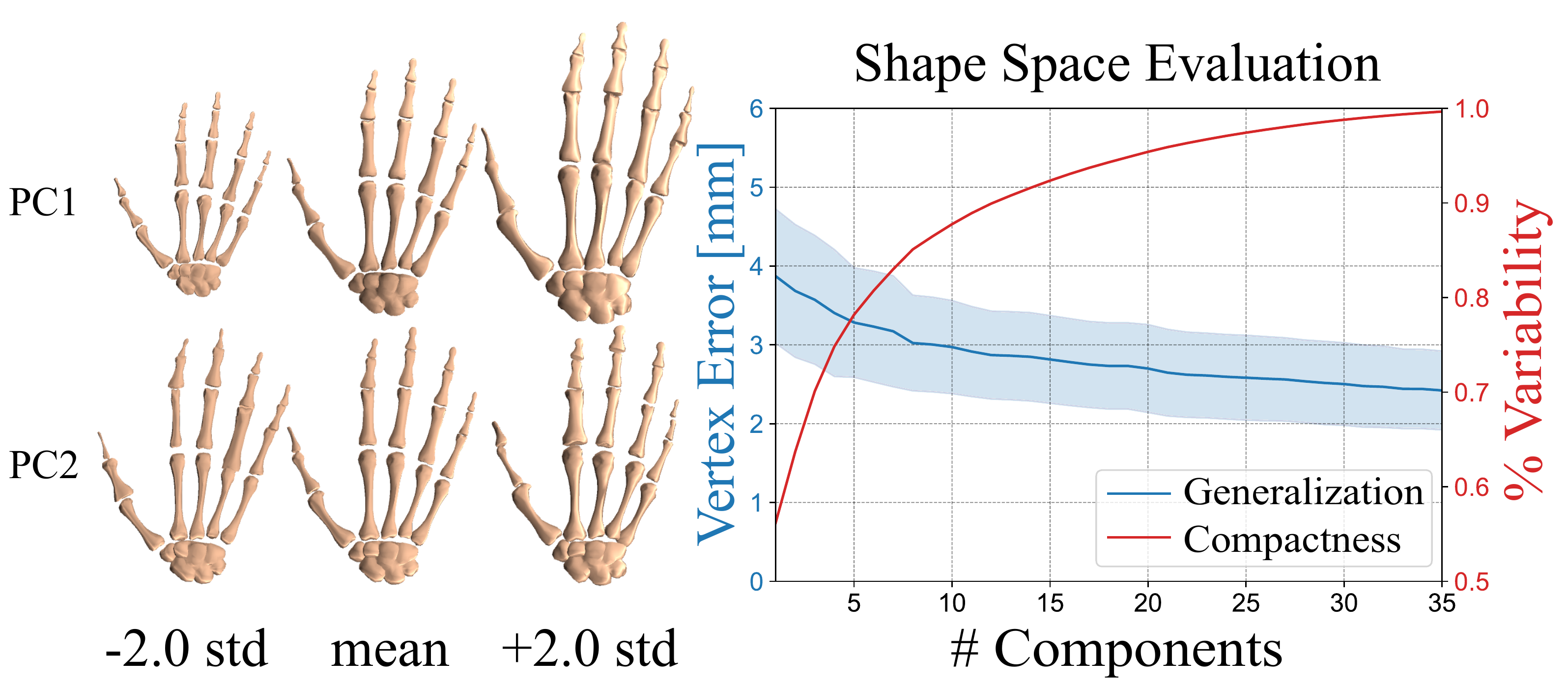}
    \caption{Left: visualization of shape space projected to the first two PCs. 
    Right: generalization and compactness curve of PIANO. We plot the mean and standard deviation of the vertex mean squared error for generalization and the percent of variability recovered with various number of PCs for compactness.}
    \label{fig:shapspace}
\end{figure}

\noindent\textbf{Anatomic inference from MRI.}
\label{sec:mm:mrinet}
As illustrated in Fig.\ref{fig:mri_net}, we propose a novel network structure for fine-grained MRI hand bone segmentation with PIANO layer.
Our encoder consists of ResNet3D and three regression branches. Each has two fully connected layers with hidden size 512 and ReLU as activation function after the first layer. The encoder extracts a latent feature from an MRI volume. Then, the extracted feature is passed to regression branches to infer pose $\theta$ and shape $\beta$ of PIANO as well as global rotation and translation $\mathcal{G}$.
PIANO layer then outputs a fine-graind semantic bone mesh.
The bone mesh is then voxelized to a regular volumetric grid with scalar values representing semantic bone label.
The SegNet takes the concatenation of the voxelized mesh and original volume as input and outputs label probabilities.
We train this network on our training set, which contains annotation of 3D joints and fine-grained bone mask. Thus the loss term can be defined as the summation of fine-grained semantic segmentation loss and PIANO loss, where the segmentation loss is the linear combination of multi-class binary cross entropy loss (BCE) and dice loss (DC):
\begin{equation}
   \mathcal{L}_{BCE} = -\frac{1}{K}\sum_{l=1}^K (y_l \log \hat{y}_l+(1-y_l) \log(1-\hat{y}_l)); 
\end{equation}
\begin{equation}
    \mathcal{L}_{DC} =  1 - \frac{2}{K}\sum_{l=1}^K \frac{y_l \hat{y}_l}{y_l + \hat{y}_l}, 
\end{equation}
where $\hat{y}$ is the sigmoid output of the network, $y$ is the one hot encoding of the ground truth segmentation. $K=20$ is number of fine-grained classes.

\noindent\textbf{Anatomic inference from RGB.}
\label{sec:mm:rgbnet}
To extend the application to RGB images, similar to the MRI-based inference, we build a network with an image encoder and two regression branches that infer pose and shape on the FreiHand~\cite{zimmermann2019freihand} with 3D joint supervision.
To account for the mismatch of joint positions, we add an additional linear layer to map from our joint to the dataset joint annotation. 
Since we only have sparse joint supervision for this task, following \cite{hasson2019learning}, we use a regularizer on the hand shape to prevent extreme mesh deformations and formulate it as  $\mathcal{L}_{\boldsymbol{\beta}} = ||\boldsymbol{\beta}||^2$, which constraints the hand shape to be close to the average shape in our MRI dataset, thus the loss is defined as $\mathcal{L}_J + \mathcal{L}_{\boldsymbol{\beta}}$.

    \begin{figure}[t]
    \centering
    \includegraphics[width=0.8\linewidth]{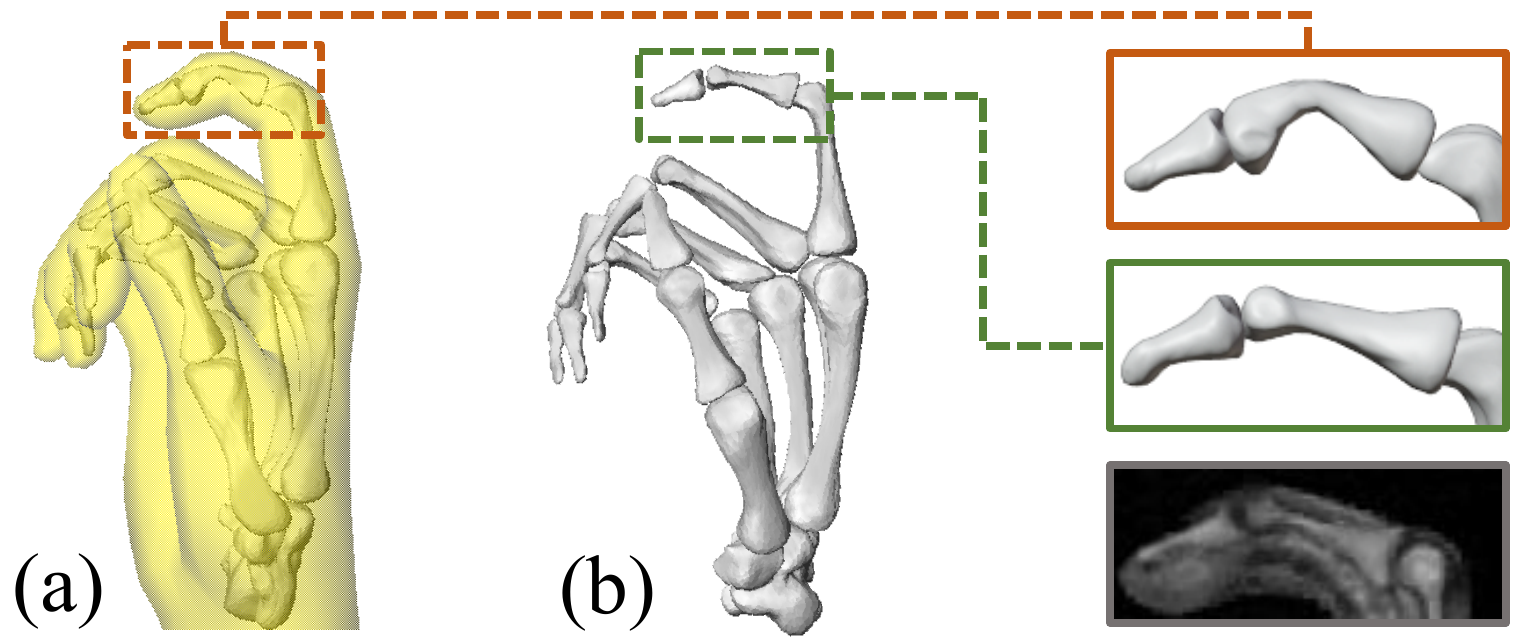}
    \caption{Deformation comparison between MANO ana PIANO. 
    The interior bone suffers from implausible bending in (a) MANO, while our model is able to maintain the bone rigidness in (b) PIANO. The bottom right is a real MRI slice of the corresponding pose.}
    \label{fig:manofail}
\end{figure}

\begin{table}[]
\centering
\begin{tabular}{c|c|c}
\hline
Methods & \textbf{PCK$\uparrow$} & \textbf{MSE~(mm) $\downarrow$}         \\ \hline
FetalPose    &  0.621     &    18.529 \\ \hline
MANO   & 0.777 & 11.167 \\ \hline
PIANO    &  \textbf{0.907}  &  \textbf{4.607} \\  \hline
\end{tabular}
\caption{Comparison with non-model based method and MANO in terms of hand joint detection error from MRI scans. We use 50 mm as the up threshold for the PCK metric.}
\label{tab:joint_err}
\end{table}

\section{Experiment}
In the following section, we present quantitative and qualitative evaluation of PIANO and the corresponding applications. 

\begin{figure*}[t]
    \centering
    \includegraphics[width=\linewidth]{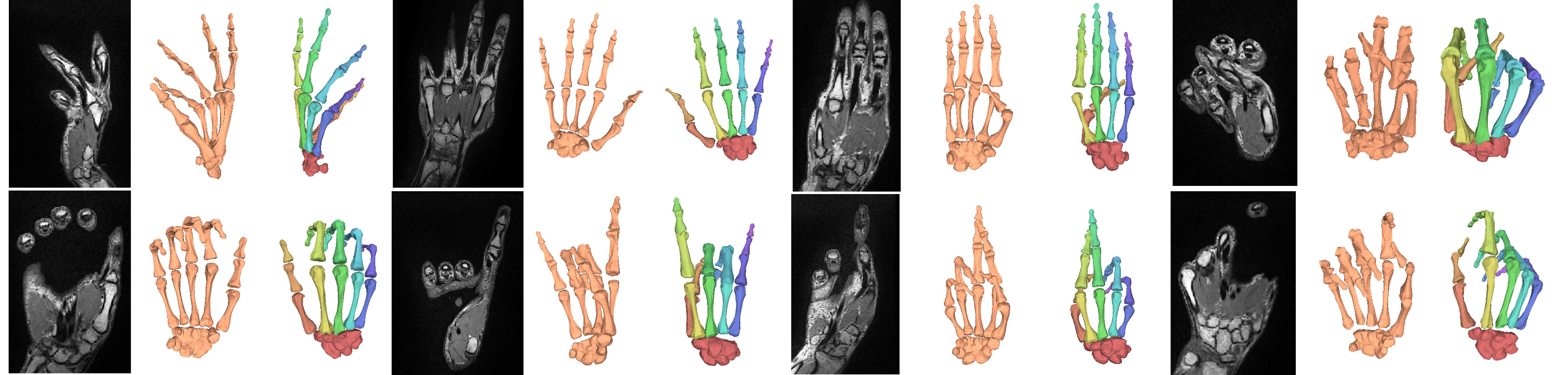}
    \caption{Results on fine-grained anatomic and semantic understanding of human hands from MRI scans.}
    \label{fig:Gallery}
\end{figure*}

\begin{figure}[t]
    \centering
    \includegraphics[width=\linewidth]{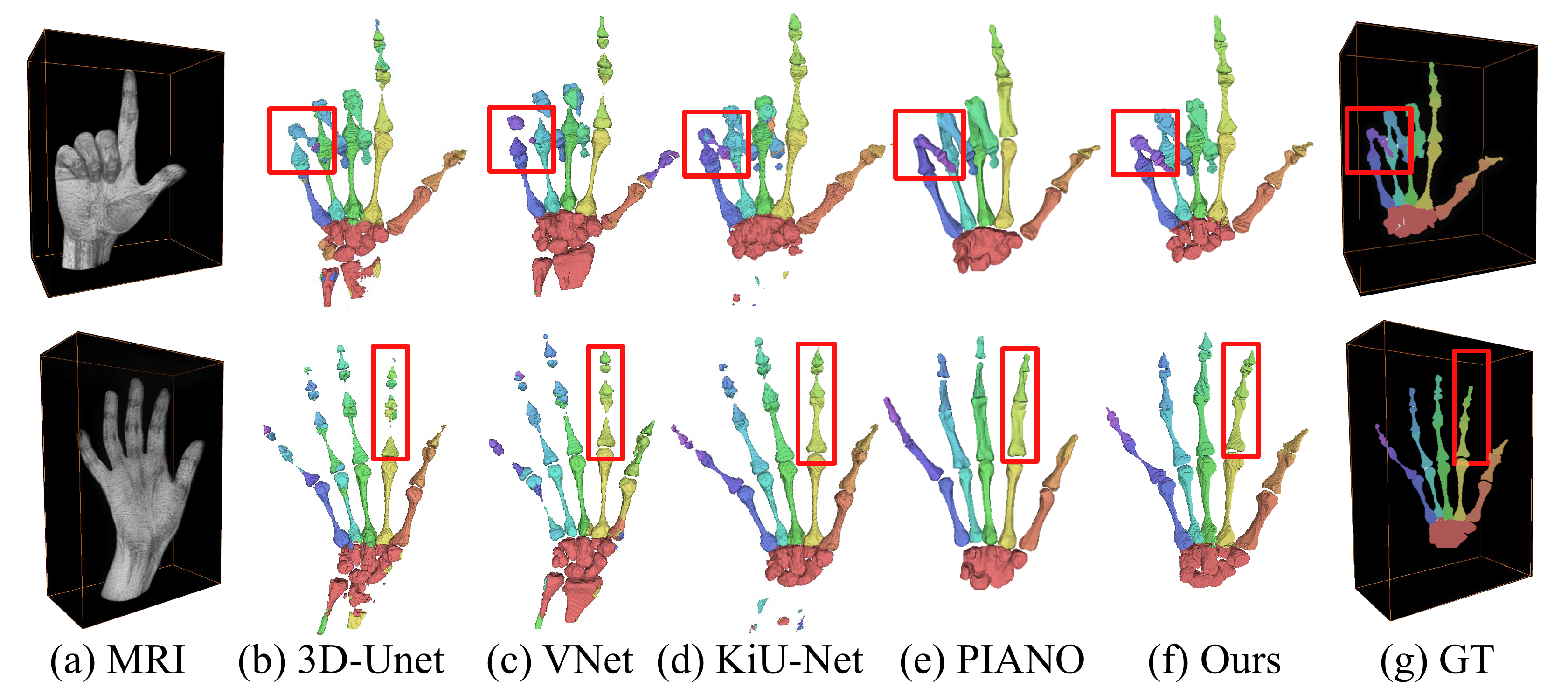}
\caption{Comparison on fine-grained MRI bone segmentation. (a) Input volume. (b, c, d) Results from 3D-Unet, VNet and KiU-Net. (e) PIANO fitting results. (f) Our network results. (g) Ground truth.}
\label{fig:segexp}
\end{figure}

\subsection{PIANO Evaluation}
\label{sec:exp_compact}
We first evaluate the shape space of PIANO qualitatively on the left of Fig.\ref{fig:shapspace}, which provides a visualization of our shape space projected
to the first two principal components (PCs).
The effect of each PC is illustrated by adding $\pm$2 standard deviations(std) to the mean shape in the center, which depicts that the first component restricts the hand size, while the second component controls the bone thickness. 
We further evaluate the compactness and generalization of our model quantitatively on the right of Fig.\ref{fig:shapspace}.

\noindent\textbf{Compactness.}
The red curve in Fig.~\ref{fig:shapspace} illustrates the compactness of our model shape space. It shows the shape variance of our dataset recovered by a varying number of components.
Note that the first principal component covers over $50\%$ of the shape space. 
Meanwhile, 10 and 20 components manage to cover $90\%$ and $95\%$ of the complete shape space.  

\noindent\textbf{Generalization.}
The blue curve in Fig.~\ref{fig:shapspace} illustrates the generalization of our model. 
We use the leave-one-out evaluation of the shapes in our dataset and report the mean squared vertex error. 
Such quantitative evaluation depicts the generalization ability of our model to unseen shapes. 
As the number of principal components increases, the mean error decreases to a minimum error of 2.5 mm achieved by the full space. 

\begin{figure}[t]
    \centering
    \includegraphics[width=\linewidth]{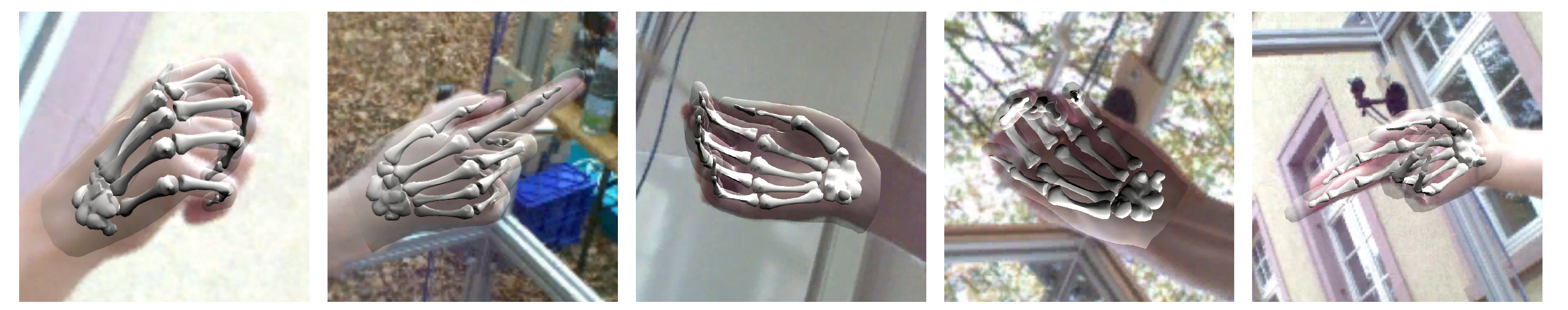}
    \caption{Qualitative results of RGB-based hand bone analysis.}
    \label{fig:rgbexp}
\end{figure}

\subsection{Comparison} \label{sec:exp_applications}
Here we compare our model on various internal hand bone inference tasks.
For the MRI-based tasks, we separate our MRI dataset into 140 training volumes and 60 testing volumes, where testing volumes contain unseen hand shapes and poses. 
We train our approach with the PIANO model only on the training set and evaluate on the test set.

\noindent\textbf{Compare with MANO.}
We compare the joint estimation ability of our PIANO model against MANO~\cite{romero2017embodied} by training both models using the encoder network proposed in Sec.\ref{sec:mm:mrinet} with joint loss and shape regularization loss. 
Note that we adopt full dimension for pose parameter and 10 dimension for shape parameters, and use Probability of Correct Keypoint ({PCK}) and Mean Squared Error ({MSE}) to measure the joint detection error.
As shown in Tab.~\ref{tab:joint_err}, our model achieves more accurate results consistently on both metrics due to our anatomically precise modeling ability.
Fig.~\ref{fig:manofail} further reveals that MANO suffers from impractical bone deformations since it is based on outer surface only.
In contrast, our PIANO model achieves more physically plausible deformation with internal modeling.
\textcolor{black}{For a more thorough comparison, we also compare against the non-model based method FetalPose~\cite{xu2019fetal} on MRI joint estimation and report the results in Tab.~\ref{tab:joint_err}. It can be seen that model based methods out performs FetalPose on both metrics. }

\noindent\textbf{Fine-grained MRI Bone Segmentation.}
We adopt PIANO and the network in Fig.~\ref{fig:mri_net} for MRI hand bone segmentation and Fig.~\ref{fig:Gallery} shows our results on various challenging hand poses.
We compare with fully supervised medical segmentation baselines
{VNet}~\cite{milletari2016v}, {3D-Unet}~\cite{cciccek20163d} and {KiU-Net}~\cite{valanarasu2020kiu}.
For quantitative analysis, the dice score ({DICE}) and the symmetric hausdorff distance ({Hausdorff}) are adopted as metrics.
As shown in Fig.~\ref{fig:segexp} and Tab.~\ref{tab:seg_dict}, the baselines suffer from severe local inference noise, while the approach using our PIANO model provides more complete results, significantly improving the performance of fine-grained bone segmentation.

\begin{table}[]
\centering
\begin{tabular}{c|c|c}
\hline
Methods & \textbf{DICE$\uparrow$}       & \textbf{Hausdorff~(mm)$\downarrow$}         \\ \hline
VNet    & 0.625 & 27.539  \\\hline
3D-UNet & 0.663 & 23.636\\\hline
KiU-Net & 0.695 & 22.839 \\\hline
Ours           & \textbf{0.717}    &  \textbf{4.02}    \\ \hline
\end{tabular}
 \caption{Comparison of fine-grained MRI-based bone segmentation.}
 \label{tab:seg_dict}
\end{table}

\noindent\textbf{Anatomic Inference from RGB.}
In Fig.~\ref{fig:rgbexp}, we show the qualitative results for obtaining PIANO model from a single RGB image, which is taken from the evaluation set of the FreiHand dataset.
Note that our PIANO results are overlayed with the MANO surface in Fig.~\ref{fig:rgbexp}. 
While MANO captures the outer hand surface, our model is able to estimate inner bone structure which is anatomically precise.

    \section{Conclusion}
We have presented PIANO, the first parametric hand bone model which enables anatomically precise modeling of the internal kinematic structure of human hands in a data-driven manner.
Our model is biologically correct, simple to animate and differentiable, based on a new dataset of high-quality MRI hand scans of various subjects and poses.
For model creation, our multi-stage registration approach and the bone-aware model learning strategy enable us to utilize the joint and bone segmentation annotations in MRI data provided by radiologists.
Furthermore, we demonstrated that our model enables more anatomically precise personalized rigging and fine-grained hand bone segmentation from an MRI scan or even a single RGB image in a data-driven fashion.
We make our model publicly available to stimulate further work and potentially benefit virtual hands in games/film, robotics, human healthcare and medical education.
In future, we plan to extend our parametric model to muscles, tendons and fats using our MRI dataset for more ingenious hand modeling.

\end{CJK}

\section*{Acknowledgments}
This work was supported by NSFC programs (61976138, 61977047), the National Key Research and Development Program (2018YFB2100500), STCSM (2015F0203-000-06) and SHMEC (2019-01-07-00-01-E00003).
We would like to thank Hui Yang, Xiaozhao Liu, Xin Tang, Zhan Jin, Guohao Jiang, Lang Mei, Lu Zou, Boliang Yu, and Yongshen Li for the help of data acquisition. 

% \appendix

\small
\bibliographystyle{named}
\bibliography{ijcai21}

\end{document}